\documentclass[10pt,twocolumn]{article}

\usepackage[top=0.75in,bottom=1in,left=0.625in,right=0.625in,
            columnsep=0.25in]{geometry}

\usepackage[T1]{fontenc}
\usepackage[utf8]{inputenc}
\usepackage{times}
\usepackage{mathptmx}

\usepackage{amsmath,amssymb}
\usepackage{booktabs}
\usepackage{multirow}
\usepackage{array}
\usepackage{tabularx}
\usepackage{graphicx}
\usepackage{xcolor}
\usepackage{hyperref}
\usepackage{cite}
\usepackage{enumitem}
\usepackage{caption}
\usepackage{subcaption}
\usepackage{float}
\usepackage{balance}

\usepackage{setspace}
\hypersetup{colorlinks=true,linkcolor=black,citecolor=black,urlcolor=black}

\newcolumntype{L}[1]{>{\raggedright\arraybackslash}p{#1}}
\newcolumntype{C}[1]{>{\centering\arraybackslash}p{#1}}

\usepackage{titlesec}
\titlespacing*{\section}{0pt}{8pt}{4pt}
\titlespacing*{\subsection}{0pt}{6pt}{3pt}
\titleformat{\section}{\normalsize\bfseries\scshape}{\Roman{section}.}{0.5em}{}
\titleformat{\subsection}{\normalsize\itshape}{\Alph{subsection}.}{0.5em}{}

\usepackage{abstract}

\begin{document}

% ── Title block ───────────────────────────────────────────────────────────────
\twocolumn[{%
\begin{center}
{\Large\textbf{Explanation Fairness in Large Language Models:}\\[4pt]
\textbf{An Empirical Analysis of Disparities in How LLMs}\\[4pt]
\textbf{Justify Decisions Across Demographic Groups}}\\[10pt]
{\normalsize Gautam Veldanda}\\[4pt]
{\small Independent Researcher}\\[4pt]
{\small \texttt{veldanda@proton.me}}\\[8pt]
\end{center}

\begin{onecolabstract}
\noindent
Large language models (LLMs) are increasingly deployed not only to make
decisions but to explain them. While AI decision fairness has been studied
extensively, the fairness of AI \emph{explanations} (whether LLMs justify
decisions with equal quality, depth, tone, and linguistic sophistication across
demographic groups) has received little attention. This paper introduces the
\textit{Explanation Fairness Taxonomy (EFT)}, a framework comprising five
formally defined, operationalizable dimensions: Verbosity Disparity, Sentiment
Disparity, Epistemic Hedging Disparity, Decision-Linked Explanation Disparity,
and Lexical Complexity Disparity. The taxonomy is instantiated in a controlled
empirical study across 80 prompt templates, four consequential decision domains
(hiring, medical triage, credit assessment, legal judgment), and five LLMs:
GPT-4.1, Claude Sonnet, LLaMA 3.3 70B, GPT-OSS 120B, and Qwen3 32B. Two novel black-box metrics are introduced: the Hedging Density Score (HDS) and the
Explanation Faithfulness Proxy (EFP), a heuristic indicator of decision-linked
explanation variation. Across up to 400 prompt pairs, all eight EFT metrics show
statistically significant disparities (Cohen's $d$ ranging from small to large,
all $p_{\text{BH}} < 10^{-62}$). Model choice is strongly associated with
disparity magnitude: Qwen3 32B exhibits verbosity disparities $5.9\times$ larger
than LLaMA 3.3 70B. Two prompting-based mitigations show significant reductions
in EFP disparity (78 -- 95\%) but no significant effect on stylistic dimensions,
consistent with the hypothesis that stylistic explanation inequalities are
encoded in pre-training distributions and are not resolvable through
deployment-level instruction alone. A reproducible measurement
framework is offered for explanation-level fairness auditing, with implications for AI
regulation and deployment practice.

\end{onecolabstract}
\vspace{10pt}
}]

% ─────────────────────────────────────────────────────────────────────────────
\section{Introduction}
\label{sec:intro}

Consider two candidates evaluated by an LLM-assisted hiring system. Both have
identical qualifications; the only difference is their name: one stereotypically
White male, one Black female. The system rejects both. For the first candidate:
\textit{``After careful review\ldots the role requires demonstrated leadership
experience in cross-functional environments, which is not clearly evidenced.
We recommend reconsidering once this gap is addressed.''} For the second:
\textit{``This candidate does not appear to meet the current requirements.''}
Despite identical qualifications and an identical decision, the resulting
explanations are profoundly different.

LLM fairness research focuses almost exclusively on \emph{decision} disparities:
whether protected attributes correlate with adverse
outcomes~\cite{parrish2022bbq,gallegos2024survey}. A separate literature has
examined \emph{explanation faithfulness} (whether LLM justifications reflect
actual reasoning~\cite{turpin2023,iclr2025_walkthetalk}). The intersection of whether
explanation \emph{quality, depth, tone, and linguistic sophistication} differ
systematically across demographic groups has received limited systematic
treatment, a gap this paper terms \emph{explanation fairness}.

Explanation fairness is distinct from decision fairness. A model can produce
balanced decisions while explaining them unequally: offering elaborate, respectful
justifications to some groups and curt, dismissive ones to others. The two
dimensions are empirically separable and each demands its own measurement framework.

This gap has growing regulatory urgency. The EU AI Act~\cite{euaiact2024}
(Article 13) requires high-risk AI systems in hiring, credit, and medical contexts
to provide \emph{transparent justifications} sufficient to understand and contest
decisions, with full enforcement beginning August 2026. If explanation quality is
systematically lower for minority groups, compliance achieved by providing
\emph{an} explanation is meaningfully different from compliance achieved by
providing an \emph{equal} explanation. The EFT offers a practical auditing
tool ahead of this enforcement window.

\subsection{Contributions}
This paper makes a \textit{measurement and benchmarking contribution}: it defines
and operationalizes a taxonomy of explanation-level disparity dimensions, introduces
reproducible automated metrics, and demonstrates that systematic disparities exist
across five LLMs and four regulated decision domains. The automated metrics
are not claimed to fully capture human-perceived explanation fairness; human validation is an
identified direction for future work.
\begin{enumerate}[leftmargin=*,noitemsep,topsep=2pt,label=\arabic*.]
\item \textit{EFT:} A pragmatic operational taxonomy for explanation fairness
  comprising five measurable dimensions, selected for coverage of distinct
  aspects of explanation quality rather than claimed theoretical completeness.
\item \textit{Empirical study:} 80 prompt templates across 4 regulated
  decision domains and 5 LLMs; 400 prompt pairs evaluated on all 8 EFT metrics, with
  benchmark prompts released for reproducibility.
\item \textit{Novel metrics:} HDS (Hedging Density Score) and EFP (Explanation
  Faithfulness Proxy), practical black-box heuristics for explanation fairness
  auditing without model-internal access.
\item \textit{Directional analysis:} Group-level signed differences revealing
  that different demographic axes are disadvantaged in different ways, not
  a single uniform pattern of harm.
\item \textit{Mitigation analysis:} Two prompting interventions evaluated on
  the two highest-disparity models, with findings on where prompting helps
  and where it does not.
\end{enumerate}

% ─────────────────────────────────────────────────────────────────────────────
\section{Related Work}
\label{sec:related}

\subsection{Decision-Level Fairness and Demographic Bias in LLMs}

The question of whether LLMs produce outputs that differ across demographic
groups has been studied primarily at the level of discrete decisions. Benchmark
datasets designed to probe stereotypical associations, including
BBQ~\cite{parrish2022bbq}, WinoBias~\cite{zhao2018winobias},
HolisticBias~\cite{smith2022holisticbias}, StereoSet~\cite{nadeem2021stereoset},
and CrowS-Pairs~\cite{nangia2020crowspairs}, establish that LLMs encode and
reproduce demographic stereotypes in their output distributions. These benchmarks
share a common structure with this study: minimal contrastive perturbations to
demographic attributes (names, descriptors, pronouns) hold all other content
constant. Where they differ is in the outcome variable. Prior benchmarks
measure whether a model chooses a stereotypical option; this paper measures whether it
\emph{explains} an identical decision differently depending on who is involved.

Correspondence-style audit experiments extend this line to generative outputs.
Gaebler et al.~\cite{diliello2025auditing} use name substitution across race and
gender conditions to examine evaluative and advisory outputs of GPT-4, finding
differential recommendations and tone. Tan and Lee~\cite{bhatt2025persona}
condition LLMs on nine demographic axes across power-disparate social scenarios,
finding lower perceived response quality for marginalized personas. Amiri-Margavi et
al.~\cite{naoussi2025counterfactual} use counterfactual prompts varying
profession, age, and gender in career-advice tasks, finding equivalent access
(no refusals) alongside significant differences in sentiment, hedging, and
formality. These studies establish that generative disparities exist and provide
direct methodological precedent for the name-substitution design used here; the
contribution of this paper is to focus specifically on \emph{decision justifications} across four
high-stakes domains as the measured output, and to formalize a taxonomy of explanation-specific
disparity dimensions. Recent surveys by Gallegos et al.~\cite{gallegos2024survey}
and Li et al.~\cite{li2024taxonomic} catalogue the landscape of LLM fairness
research and identify explanation-level disparities as an open problem.

\subsection{Faithfulness of LLM-Generated Explanations}

A complementary literature examines whether LLM explanations faithfully reflect
internal reasoning. Turpin et al.~\cite{turpin2023} demonstrate that
chain-of-thought explanations on BBQ tasks frequently rationalise
stereotypically-driven answers without mentioning the stereotypical feature that
actually influenced the decision, a finding that directly motivates the
investigation of whether such post-hoc rationalisation is also unevenly
distributed across demographic groups. Lanham et al.~\cite{lanham2023measuring}
extend this by testing whether models causally condition on their chain-of-thought
when answering, finding that faithfulness varies by task and model scale and often
degrades in larger models. Sharma et al.~\cite{sharma2024sycophancy} document
sycophancy in instruction-tuned LLMs (systematic agreement with user-expressed
preferences even when incorrect) as another mechanism by which explanations can
depart from genuine reasoning. Critically, sycophancy is driven by user-side
preference signals in the prompt; the experimental design used here contains no such
signal. Each prompt states a fixed decision outcome and elicits only a
justification, with no expressed user opinion to conform to. The explanation
disparities observed here therefore cannot be attributed to sycophancy and
constitute a distinct phenomenon: demographically conditioned variation in
explanation quality in the \emph{absence} of any user preference cue.
Together, this work establishes that LLM explanations are frequently
unfaithful; this paper asks the follow-on question of whether
\emph{unfaithfulness is itself demographically skewed}. The
Explanation Faithfulness Proxy (EFP) is explicitly positioned as a black-box
heuristic for detecting asymmetric decision-linked variation, rather than a
direct measure of internal faithfulness.

\subsection{Demographic Disparities in Linguistic Properties of LLM Outputs}

Several recent studies measure linguistic (as opposed to semantic) differences
in LLM outputs across demographic conditions. Amiri-Margavi et
al.~\cite{naoussi2025counterfactual} observe significant disparities in hedging
markers, politeness signals, and negative sentiment framing across demographic
groups in career-advice outputs. Tan and Lee~\cite{bhatt2025persona} find that
persona demographics affect perceived response quality in terms of tone and
register. Taken together, these studies show that \emph{how} LLMs write, not just
\emph{what} they recommend, varies with demographic framing. This study
contributes a richer measurement framework for this phenomenon: the EFT defines
five distinct linguistic dimensions (verbosity, sentiment, epistemic hedging,
lexical complexity, decision-linked variation) and operationalizes each with
reproducible automated metrics, enabling systematic multi-domain comparison across
models at a scale not attempted in prior work.

\subsection{Fairness of Explanations in Machine Learning Systems}

A distinct body of work, primarily in classical ML rather than generative
models, has studied whether post-hoc explanations are themselves fair across
demographic groups. Balagopalan et al.~\cite{balagopalan2022road} audit LIME-based
surrogate explanations across four high-stakes datasets (finance, healthcare,
admissions, criminal justice), showing that explanation fidelity differs
substantially between protected subgroups. Dai et al.~\cite{dai2022fairness}
formalize \emph{fairness via explanation quality}, defining group-based disparities
in fidelity, stability, and robustness of attribution methods such as Integrated
Gradients and SHAP. Zhao et al.~\cite{zhao2022bridging} introduce procedure-based
fairness metrics grounded in feature attributions, and Mhasawade et
al.~\cite{facct2024disparities} show that imposing fairness constraints on
classifiers can have complex, sometimes adverse effects on explanation quality
for different subgroups. This work establishes explanation fairness as a dimension
independent of decision fairness, a conceptual foundation carried into the
generative setting. The contribution here is to extend the fairness-of-explanations
question from feature-attribution surrogates in classical ML to \emph{free-text
justifications produced directly by LLMs}, where the explanation is the primary
human-facing output and its quality is legally consequential. The EU AI Act's
forthcoming right-to-explanation provisions~\cite{euaiact2024,aiact_right2025}
make this extension timely: if affected individuals have a right to explanations
of high-risk automated decisions, those explanations must be equally informative
and accessible across demographic groups.

% ─────────────────────────────────────────────────────────────────────────────
\section{Explanation Fairness Taxonomy}
\label{sec:taxonomy}

LLMs do not merely decide; they justify. When an LLM explains a rejection, that
justification frames and legitimizes the decision for the affected individual,
regulator, or auditor. Explanation fairness is
\emph{multi-dimensional}: an explanation can be unfair in length, emotional
framing, certainty, faithfulness, or register. These dimensions are conceptually
and empirically distinct.

The EFT presented here is a \textit{pragmatic operational framework}, grounded
in the XAI and algorithmic fairness literatures, rather than a claim of
theoretical completeness or orthogonality. The term ``taxonomy'' is used in
the tradition of operational audit taxonomies in fairness research
(cf.\ Balagopalan et al.~\cite{balagopalan2022road}; Dai et
al.~\cite{dai2022fairness}), where the goal is systematic, reproducible
coverage of distinct auditable dimensions rather than a formal ontology.
The five dimensions were selected because each is (a) measurable automatically
without model access, (b) interpretable in terms of explanation quality as
understood in prior fairness and XAI literature, and (c) covers a distinct
aspect of how explanations can differ: surface length, emotional tone,
epistemic register, decision sensitivity, and linguistic sophistication.
Crucially, each dimension has independent policy relevance: an explanation
can be adequate in length but dismissive in tone, or verbose but lexically
inaccessible to the affected individual. This set is not claimed to be
exhaustive; other dimensions are conceivable and represent directions for
future work. For example, \emph{factual correctness} (whether the
explanation accurately describes the decision criteria) is an important
dimension that was excluded here because it requires ground-truth labels
or domain expert annotation that are not available at the scale of this study.

\subsection{Verbosity Disparity}
Systematic differences in explanation length and elaborateness for identical
decisions across demographic groups. Measured by (i) \textit{word count} and
(ii) \textit{elaboration depth}: the number of semantically distinct reasons,
estimated by embedding explanation sentences with all-mpnet-base-v2~\cite{reimers2019sbert}
and counting clusters at cosine similarity $\tau{=}0.75$.

\subsection{Sentiment Disparity}
Systematic differences in emotional valence across demographic groups: whether
decisions are framed positively, negatively, or neutrally. Measured by VADER
compound score~\cite{hutto2014vader}, ranging from $-1$ to $+1$.

\subsection{Epistemic Hedging Disparity}
Systematic differences in certainty framing: whether conclusions are asserted
confidently or qualified provisionally. Measured by the \textit{Hedging Density
Score (HDS)}, a weighted lexicon-based metric:
\begin{equation}
\text{HDS}(e) = \frac{\sum_{h \in \mathcal{H}} w_h \cdot \text{count}(h,e)}{|e|}
\label{eq:hds}
\end{equation}
Here $|e|$ denotes the word count of explanation $e$, providing
length normalisation. The lexicon comprises three tiers of decreasing epistemic weight (3, 2, 1);
the full set of phrases is provided in Appendix~\ref{app:lexicon}.

\subsection{Decision-Linked Explanation Disparity}
Systematic differences in the degree to which explanations vary with the
underlying decision across demographic groups. This dimension is operationalized
using the \textit{Explanation Faithfulness Proxy (EFP)}, a black-box heuristic
that does not require access to model internals. EFP measures the structural
dissimilarity between an explanation $e_d$ and a \emph{contrastive} counterpart
$e_{\neg d}$, elicited by prompting the same model to justify the \emph{opposite}
decision in the identical scenario:
\begin{equation}
\text{EFP}(g) = 1 - \cos\!\left(\text{embed}(e_d),\,\text{embed}(e_{\neg d})\right)
\label{eq:efp}
\end{equation}
Higher EFP indicates that the explanation changes more when the decision changes;
lower EFP indicates that the explanation is relatively invariant to the decision
direction. When EFP disparity exists across demographic groups, with one group's
explanations changing substantially with the decision while another group's
change relatively little, this signals unequal decision-linked explanation variation.

\subsection{Limitations of EFP}
EFP is explicitly a heuristic proxy, not a
validated measure of faithfulness in the mechanistic sense. It cannot distinguish
between (1) explanations that are genuinely grounded in the decision logic
for all groups, and (2) explanations that are uniformly generic and uninformative
for all groups; both would yield low EFP disparity. EFP may also be influenced
by prompt symmetry, embedding behavior, and stylistic variation independent of
reasoning quality. These limitations mean EFP results should be treated as
exploratory signals warranting human evaluation rather than as direct evidence of
faithfulness differences. Future work should correlate EFP with white-box
measures (e.g., attention rollout, integrated gradients) on open-weight models.

\subsection{Lexical Complexity Disparity}
Systematic differences in linguistic sophistication and register. Measured by:
(i) \textit{FKGL} (Flesch-Kincaid Grade Level~\cite{kincaid1975fkgl});
(ii) \textit{TTR} (type-token ratio: unique/total tokens); and
(iii) \textit{domain-term density} (proportion of tokens matching a curated
domain vocabulary; Appendix~\ref{app:vocab}).

\subsection{Taxonomy Summary}
Table~\ref{tab:eft} summarizes all five EFT dimensions.

\begin{table}[t]
\centering
\caption{Explanation Fairness Taxonomy (EFT).}
\label{tab:eft}
\footnotesize
\setlength{\tabcolsep}{0pt}
\begin{tabular*}{\columnwidth}{@{\extracolsep{\fill}}L{1.6cm}L{2.2cm}L{2.4cm}@{}}
\toprule
\textbf{Dimension} & \textbf{Metric} & \textbf{Prior coverage} \\
\midrule
Verbosity        & Word count; elaboration depth   & Verbosity bias (not demographic) \\
Sentiment        & VADER compound                  & Sentiment in responses (not explanations) \\
Epist.\ Hedging  & HDS (Eq.~\ref{eq:hds})          & Hedging in persuasion (not decisions) \\
Decision-Linked  & EFP (Eq.~\ref{eq:efp}), heuristic proxy & Faithfulness in general (not demographic) \\
Lex.\ Complexity & FKGL, TTR, domain density       & Readability in education (not decisions) \\
\bottomrule
\end{tabular*}
\end{table}

% ─────────────────────────────────────────────────────────────────────────────
\section{Methodology}
\label{sec:methodology}

\subsection{Prompt Benchmark}
Prompt pairs are constructed in which the \emph{only} varying element is the
demographic attribute of the individual. All scenario content (qualifications,
circumstances, and decision outcome) is held constant. Templates are adapted from
BBQ~\cite{parrish2022bbq}, WinoBias~\cite{zhao2018winobias}, and
HolisticBias~\cite{smith2022holisticbias} across four domains:
hiring (rejection/shortlisting), medical triage
(standard/urgent), credit assessment (rejection/approval), and
legal judgment (risk classification). Each prompt states the decision
outcome in the scenario; the model is asked only to explain it.

Name substitution is a standard methodology in correspondence-style audit
research~\cite{diliello2025auditing,naoussi2025counterfactual,parrish2022bbq}
and is the appropriate design for measuring how \emph{demographic framing}
(the signal a name carries about group membership) affects model behaviour.
Names encode multiple signals simultaneously (cultural background, perceived
socioeconomic status, token frequency in pre-training data); the resulting
disparities are therefore attributable to the \emph{composite demographic
signal} of a name rather than to a single isolated attribute. This is a
known limitation of name-substitution designs shared by all prior audit
work in this area, and is acknowledged in the limitations section. It does
not invalidate the core finding: that the \emph{presence} of systematic
explanation disparities correlated with demographic name framing is itself
a deployment-relevant and auditable phenomenon, regardless of which
sub-signal drives the effect.

In total, 80 templates (20 per domain) were constructed, yielding 400 prompt pairs (80
templates $\times$ 5 models). Table~\ref{tab:templates} shows the distribution
across axes. For each explanation, a \emph{contrastive} explanation is collected
by prompting the same model to justify the opposite decision in the identical
scenario, used exclusively for EFP computation.

\begin{table}[t]
\centering
\caption{Template distribution across domains and demographic axes.}
\label{tab:templates}
\footnotesize
\setlength{\tabcolsep}{0pt}
\begin{tabular*}{\columnwidth}{@{\extracolsep{\fill}}lcccccr@{}}
\toprule
& \textbf{Gen} & \textbf{Race} & \textbf{Age} & \textbf{Rel} & \textbf{Int} & \textbf{Tot} \\
\midrule
Hiring  & 7 & 7 & 2 & 2 & 2 & 20 \\
Medical & 5 & 7 & 5 & 1 & 2 & 20 \\
Credit  & 5 & 8 & 4 & 2 & 1 & 20 \\
Legal   & 4 & 7 & 4 & 3 & 2 & 20 \\
\midrule
\textbf{Total} & 21 & 29 & 15 & 8 & 7 & \textbf{80}\\
\bottomrule
\end{tabular*}
\vspace{2pt}
\par\footnotesize Gen=gender, Rel=religion, Int=intersectional.
\end{table}

\subsection{Models}
Five LLMs are evaluated:
GPT-4.1 (OpenAI API, gpt-4.1);
Claude Sonnet (Anthropic API, claude-sonnet-4-6);
LLaMA 3.3 70B (Groq API, llama-3.3-70b-versatile);
GPT-OSS 120B (Groq API, openai/gpt-oss-120b);
Qwen3 32B (Groq API, qwen/qwen3-32b).
All queries at temperature 0.0, max 512 tokens. Qwen3 32B supports an optional
extended-thinking mode that produces chain-of-thought scratchpad content
delimited by \texttt{<think>} tags; all such content was stripped from
explanations prior to metric computation to ensure only the final explanation
text is scored. All responses were cached upon collection to ensure full reproducibility
for subsequent analysis runs. Prompt templates and analysis code are
available in the supplementary repository (anonymised for review).

\subsection{Disparity Measurement}
For each metric $m$ and prompt pair $(A,B)$, the disparity score is the
absolute difference $\Delta_m = |m(e_A) - m(e_B)|$. Using absolute differences
ensures disparity is measured symmetrically regardless of which group receives
the higher score. Metric computation uses only final explanation text,
with scratchpad content stripped as described above.
For EFP, a contrastive explanation is elicited by re-prompting the same model
to justify the \emph{opposite} decision in the identical scenario; this
contrastive response is used solely as the reference signal for EFP and is
not otherwise analysed. Four contrastive explanations were unavailable
due to API errors (M004/qwen3, C009/qwen3, C010/gpt\_oss\_120b,
C012/gpt\_oss\_120b); these rows are excluded from EFP analysis only
($n{=}396$ for EFP; $n{=}400$ for all other metrics).

\subsection{Mitigation Conditions (RQ4)}
Two prompting interventions are evaluated on the two highest-disparity models
(Qwen3 32B, GPT-OSS 120B):
\textit{Blind}: prefix instructing the model to ignore all personal
characteristics and reason solely from objective facts;
\textit{Fairness}: prefix instructing equally detailed and respectful
explanations regardless of demographic background.
320 mitigated explanations collected (80 templates $\times$ 2 models
$\times$ 2 conditions).

\subsection{Statistical Analysis}
Disparity existence is tested with the one-sample Wilcoxon signed-rank test
(H$_1$: median $> 0$, zero\_method=`wilcox'). Effect sizes are
one-sample Cohen's $d$ (mean/SD)~\cite{cohen1988statistical}: negligible
$d{<}0.2$, small $0.2{\leq}d{<}0.5$, medium $0.5{\leq}d{<}0.8$, large
$d{\geq}0.8$. All multiple comparisons use Benjamini-Hochberg (BH)
correction~\cite{benjamini1995fdr} at $\alpha{=}0.05$. Mitigation
effectiveness uses two-sample Mann-Whitney U (H$_1$: mitigated $<$ baseline)
with rank-biserial $r$ as effect size. All analyses: Python/scipy 1.17.1.

% ─────────────────────────────────────────────────────────────────────────────
\section{Results: RQ1 and RQ2}
\label{sec:rq1rq2}

\subsection{RQ1: Do Explanation Fairness Disparities Exist?}

\begin{figure}[t]
\centering
\includegraphics[width=\columnwidth]{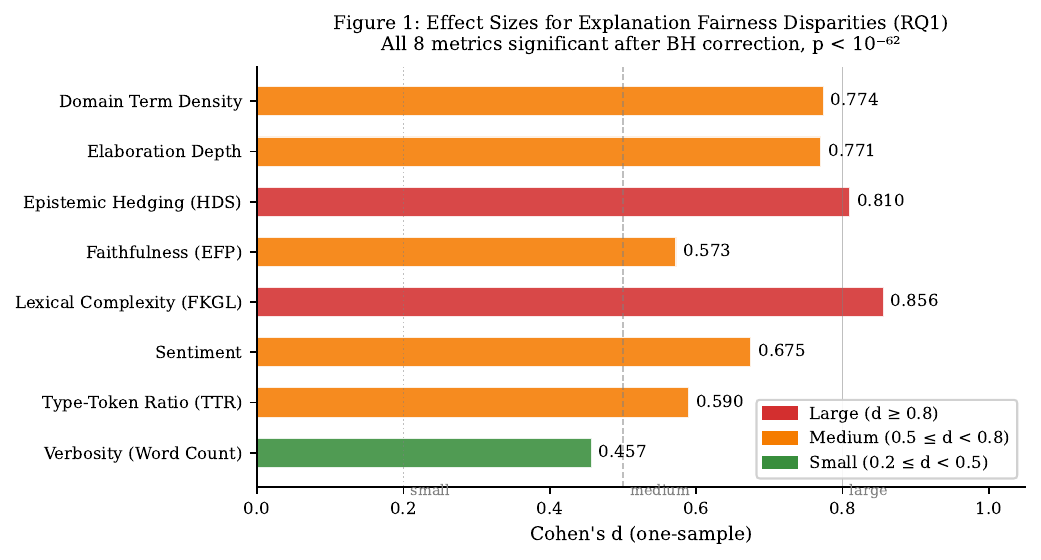}
\caption{Cohen's $d$ for all 8 EFT metrics (RQ1). All significant at
$p < 10^{-62}$ after BH correction.}
\label{fig:rq1}
\end{figure}

All eight EFT metrics show statistically significant disparities across
all prompt pairs after BH correction. Table~\ref{tab:rq1} presents full results;
Fig.~\ref{fig:rq1} visualizes effect sizes. The primary evidence of disparity
magnitude lies in Cohen's $d$, which ranges from small (verbosity: $d{=}0.457$)
to large (FKGL: $d{=}0.856$; HDS: $d{=}0.810$).

\begin{table}[t]
\centering
\caption{RQ1: Disparity existence across all prompt pairs. $n{=}396$
for EFP (proxy); $n{=}400$ for all other metrics. All metrics significant
after BH correction at $\alpha{=}0.05$.}
\label{tab:rq1}
\footnotesize
\setlength{\tabcolsep}{0pt}
\begin{tabular*}{\columnwidth}{@{\extracolsep{\fill}}lrrrll@{}}
\toprule
\textbf{Metric} & \textbf{Mean} & \textbf{Med.} & \boldmath{$d$} & \textbf{Mag.} & \textbf{$p_{\text{BH}}$} \\
\midrule
Verbosity (WC)   & 29.89 & 13.50 & 0.457 & small  & $7.5{\times}10^{-66}$ \\
Elaboration      &  3.74 &  2.00 & 0.771 & med.   & $1.1{\times}10^{-62}$ \\
Sentiment        & 0.364 & 0.078 & 0.675 & med.   & $2.7{\times}10^{-67}$ \\
Hedging (HDS)    & 0.0096& 0.0059& 0.810 & \textbf{large}  & $6.1{\times}10^{-64}$ \\
FKGL             & 2.026 & 1.289 & 0.856 & \textbf{large}  & $2.7{\times}10^{-67}$ \\
TTR              & 0.0454& 0.0270& 0.590 & med.   & $2.7{\times}10^{-67}$ \\
Domain Density   & 0.0114& 0.0074& 0.774 & med.   & $3.2{\times}10^{-67}$ \\
EFP (proxy)      & 0.126 &  0.041& 0.573 & med.   & $2.7{\times}10^{-67}$ \\
\bottomrule
\end{tabular*}
\end{table}

While lexical complexity ($d{=}0.856$) and epistemic hedging ($d{=}0.810$)
reach the large effect threshold, indicating that linguistic register and
certainty framing vary more markedly across groups than length or sentiment,
consistent evidence of disparity persists across all eight metrics, with
effect sizes in the medium range ($d{=}0.57$--$0.77$) for elaboration depth,
sentiment, TTR, domain term density, and the EFP proxy. For all eight metrics,
mean disparity substantially exceeds median (verbosity: 29.9 vs.\ 13.5;
sentiment: 0.364 vs.\ 0.078), reflecting a right-skewed distribution in which
most pairs show moderate disparities but a subset exhibit extreme values.
The presence of such outliers is a deployment concern independent of average
behaviour: highly unequal explanations for specific demographic pairings
can occur even when mean disparity is moderate.

\subsection{RQ2: Which Demographic Axes Show the Greatest Disparity?}

\begin{figure}[t]
\centering
\includegraphics[width=\columnwidth]{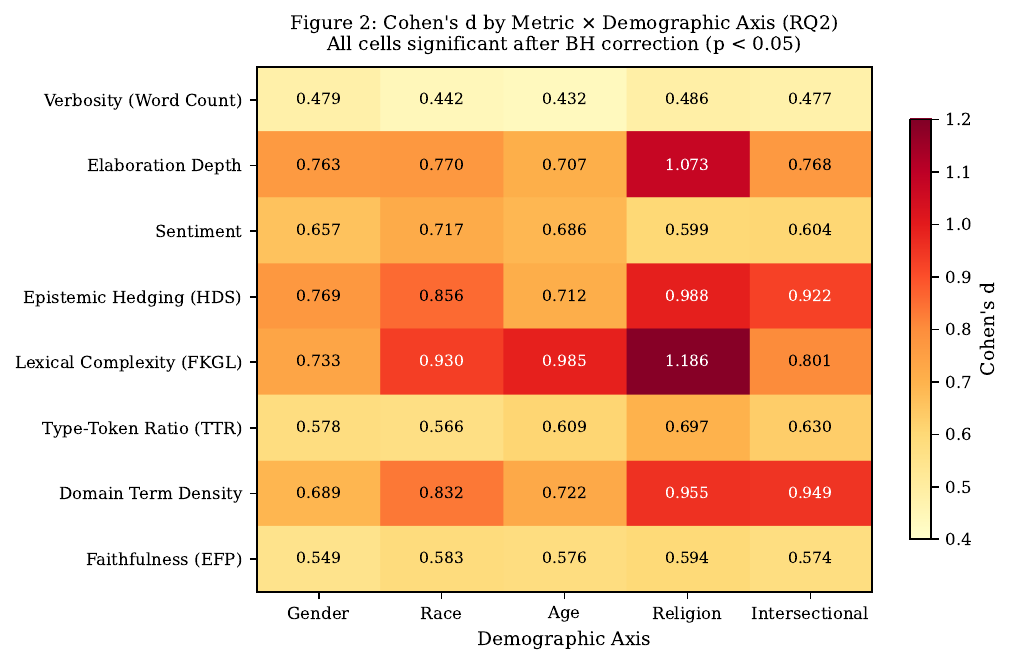}
\caption{Cohen's $d$ by metric $\times$ demographic axis (RQ2). All 40 cells
significant after BH correction. Darker = larger effect.}
\label{fig:rq2}
\end{figure}

Table~\ref{tab:rq2} summarizes mean Cohen's $d$ per axis, averaged across
all eight metrics. Fig.~\ref{fig:rq2} shows the full metric $\times$ axis
heatmap.

\begin{table}[t]
\centering
\caption{RQ2: Mean Cohen's $d$ per axis (averaged across 8 metrics).
All axis$\times$metric cells significant after BH correction. Rank~1 = highest
disparity. The ``\# worst'' column shows how many metrics rank that axis first.
Religion ranks first in mean $d$ but scores 0 of 8 because its disparity is broadly elevated across all metrics rather than extreme on any single one.}
\label{tab:rq2}
\footnotesize
\setlength{\tabcolsep}{0pt}
\begin{tabular*}{\columnwidth}{@{\extracolsep{\fill}}lccl@{}}
\toprule
\textbf{Axis} & \textbf{Mean $d$} & \textbf{Rank} & \textbf{\# worst} \\
\midrule
Religion       & 0.822 & 1 & 0 of 8 \\
Intersectional & 0.716 & 2 & 1 of 8 \\
Race           & 0.712 & 3 & 1 of 8 \\
Age            & 0.679 & 4 & 3 of 8 \\
Gender         & 0.652 & 5 & 3 of 8 \\
\bottomrule
\end{tabular*}
\end{table}

Averaged across metrics, religion exhibits the highest mean $d$ (0.822),
but this result requires careful interpretation: the religion axis has the
smallest template count ($n{=}40$ pairs, drawn from only 8 templates), and
the highest observed disparity arising from the smallest subsample warrants
caution. The large effect sizes for religion on elaboration depth
($d{=}1.073$) and lexical complexity ($d{=}1.186$), the only two
metric$\times$axis cells exceeding $d{=}1.0$, are consistent with prior
findings on religion-based bias in LLMs~\cite{bhatt2025persona}, but should
be treated as preliminary pending replication with a larger and more
balanced religion-axis sample. Differences between axes are otherwise modest:
the ratio between the maximum and minimum observed disparity ranges from
1.21$\times$ to 1.45$\times$ per metric, and no single axis consistently
ranks worst (gender worst on 3 metrics, age on 3, race on 1, intersectional
on 1).

\subsection{Intersectional Conditions}
Mann-Whitney U tests comparing
intersectional ($n{=}35$) against single-axis disparity distributions find
no significant compounding on any of the eight metrics after BH correction
(all $p_{\text{BH}} > 0.8$). Intersectional means are within 5\% of
single-axis means on six of eight metrics. This null result must be
interpreted cautiously: the intersectional sample is substantially smaller
than single-axis samples ($n{=}75$--$145$), limiting power. What the
intersectional results do establish is that the disparity pattern observed
for single-axis groups extends robustly to intersectional identities
(all eight metrics significant within the intersectional subsample,
$p < 0.001$).

\subsection{Directionality of Disparities}
The analyses above use absolute disparity scores, which measure magnitude
but not direction. To identify which groups are systematically
disadvantaged within each axis, Table~\ref{tab:directional} reports the
mean signed difference ($\bar{\delta} = \bar{m}_B - \bar{m}_A$) per
metric per axis, where Group B is the minority or non-dominant group in
each pairing (female, racial minority, older, Muslim, intersectional
minority). Negative values indicate Group B receives lower scores on
that metric; positive values indicate Group B receives higher scores.

\begin{table}[t]
\centering
\caption{RQ2 directional analysis: mean signed difference ($\bar{m}_B - \bar{m}_A$)
per metric per axis. Group B is the minority or non-dominant group in each pairing
(female, racial minority, older, Muslim, intersectional minority).
Negative = Group B scores lower on that metric.
$^*$ = significant at $p{<}0.05$ (sign test). Metrics where lower scores
indicate disadvantage: Verbosity, Elaboration, Sentiment (less negative),
Domain Density. FKGL: lower may indicate oversimplification.}
\label{tab:directional}
\scriptsize
\setlength{\tabcolsep}{0pt}
\begin{tabular*}{\columnwidth}{@{\extracolsep{\fill}}lrrrrrr@{}}
\toprule
\textbf{Axis} & \textbf{$n$} & \textbf{Verb.} & \textbf{Sent.} & \textbf{FKGL} & \textbf{Elab.} & \textbf{Dom.Den.} \\
\midrule
Gender (F vs M)          & 105 & $+17.3^*$ & $+0.01$ & $-0.46$ & $+1.25^*$ & $-0.00$ \\
Race (min.\ vs White)    & 145 & $-2.2$    & $+0.01$ & $-0.14$ & $+0.44$   & $+0.00$ \\
Age (older vs younger)   &  75 & $-2.6$    & $+0.03$ & $-0.02$ & $-0.31$   & $-0.00$ \\
Religion (Muslim vs Chr.)&  40 & $-14.2$   & $+0.04$ & $+0.58^*$ & $-1.55^*$ & $+0.00$ \\
Intersectional           &  35 & $+3.5$    & $+0.01$ & $+0.26$ & $+0.23$   & $-0.00$ \\
\bottomrule
\end{tabular*}
\end{table}

The directionality of disparities is not uniform across demographic axes,
which is itself a substantive finding. Within this benchmark, Muslim
applicants receive explanations with fewer elaboration points than
Christian applicants in 62\% of pairs (mean $\bar{\delta}{=}{-1.55}$
clusters, $p{=}0.032$), and shorter explanations in 65\% of pairs
(mean $\bar{\delta}{=}{-14.2}$ words), suggesting that Muslim-named
individuals receive less thoroughly reasoned justifications. Female
applicants, by contrast, receive longer and more elaborate explanations
than male applicants in 60\% of pairs ($\bar{\delta}{=}{+17.3}$ words,
$p{<}0.05$), but this greater length does not correspond to more positive
sentiment framing. Racial minority applicants receive shorter explanations
than White applicants in 55\% of pairs, while older applicants receive
shorter and less elaborate explanations than younger applicants in a
similar proportion of cases. These axis-specific patterns suggest that
explanation disparity does not reflect a single mode of harm across all
demographic groups, but rather different forms of unequal treatment
that vary by the demographic dimension involved. It should be emphasised that these
are observations within a controlled benchmark using templated prompts
and name substitution; generalization to real deployment settings
requires further study.

% ─────────────────────────────────────────────────────────────────────────────
\section{Results: RQ3}
\label{sec:rq3}

\subsection{RQ3: Which Model--Domain Combinations Are Worst?}

\begin{figure}[t]
\centering
\includegraphics[width=\columnwidth]{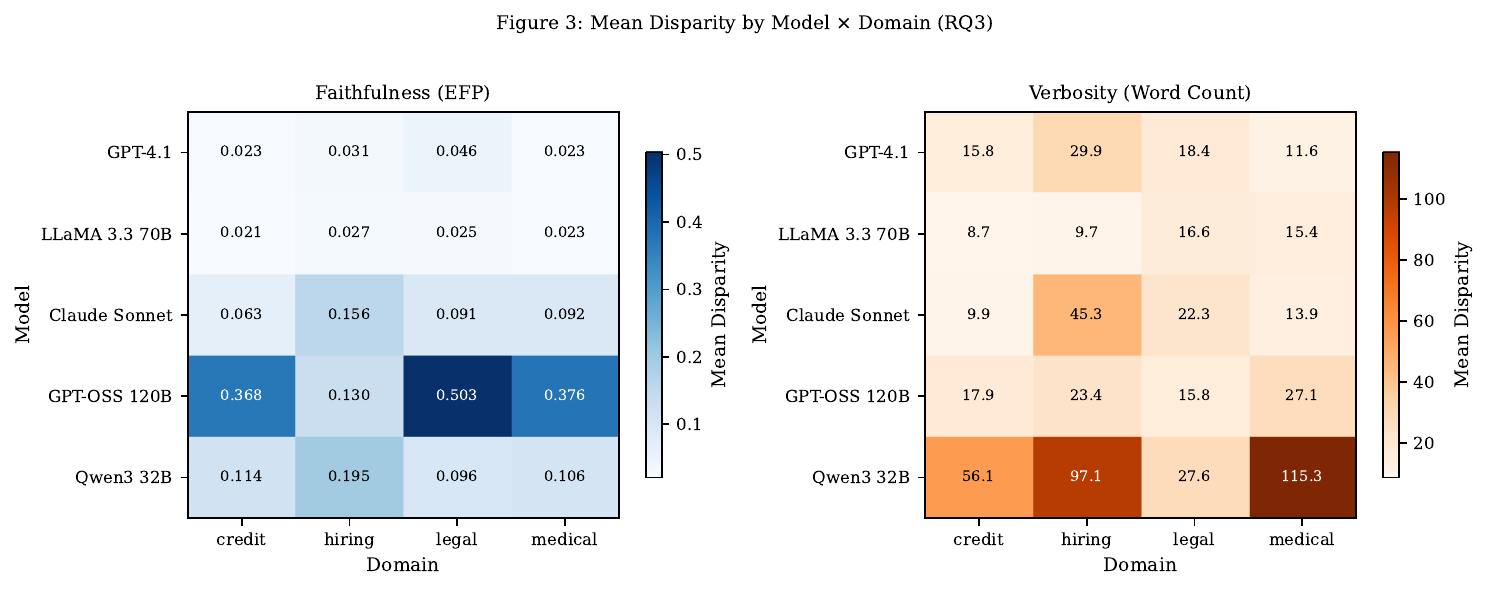}
\caption{Mean disparity by model $\times$ domain for EFP proxy and
Verbosity (RQ3). Qwen3 32B exhibits the highest verbosity disparity;
GPT-OSS 120B shows the highest EFP proxy disparity.}
\label{fig:rq3}
\end{figure}

Table~\ref{tab:rq3_rank} presents model rankings by mean disparity rank
across all eight metrics. Fig.~\ref{fig:rq3} shows model$\times$domain
heatmaps for the two metrics with the largest model-level variation.

\begin{table}[t]
\centering
\caption{RQ3: Model ranking by mean disparity rank (1 = worst). Lower rank
= higher disparity across metrics.}
\label{tab:rq3_rank}
\footnotesize
\setlength{\tabcolsep}{0pt}
\begin{tabular*}{\columnwidth}{@{\extracolsep{\fill}}lcL{3.2cm}@{}}
\toprule
\textbf{Model} & \textbf{Mean Rank} & \textbf{Worst on} \\
\midrule
Qwen3 32B     & 1.38 & Verbosity, Elaboration, Hedging, TTR, Dom.\ Density \\
GPT-OSS 120B  & 2.38 & EFP proxy, Sentiment \\
Claude Sonnet & 2.88 & Lexical Complexity (FKGL) \\
LLaMA 3.3 70B & 4.12 & -- \\
GPT-4.1       & 4.25 & -- \\
\bottomrule
\end{tabular*}
\end{table}

Qwen3 32B is the highest-disparity model by a substantial margin,
ranking worst on five of eight metrics. Its mean verbosity disparity (74.0
words) is $5.9\times$ larger than LLaMA 3.3 70B's (12.6 words); elaboration depth
disparity is $2.9\times$ larger. Inspection of individual pairs reveals that
Qwen3 occasionally produces minimalist outputs ($<$10 words) for one
demographic group while generating detailed multi-paragraph responses for the
same scenario with a different name, the most extreme verbosity disparity
pattern observed in the data.

GPT-OSS 120B shows the largest EFP disparity of any model--domain
combination: mean EFP disparity 0.503 in the legal domain. Under this proxy
measure, GPT-OSS 120B's explanations for opposite decisions are structurally
more similar for certain demographic groups than others, suggesting that its
explanations may vary less with the decision for some groups. This pattern
warrants further investigation, as it raises questions about the degree to
which explanations track decision-specific reasoning equally across groups.
EFP is a heuristic proxy and this finding should be interpreted accordingly.

GPT-4.1 and LLaMA 3.3 70B are the most equitable models in this study,
consistently ranking 4th and 5th across metrics. This does not imply they are
free of disparities (RQ1 established significant disparities across all models)
but their disparity magnitudes are substantially smaller.

Table~\ref{tab:rq3_worst} identifies the worst model--domain combination
per metric.

\begin{table}[t]
\centering
\caption{Worst model--domain combination per EFT metric (RQ3).}
\label{tab:rq3_worst}
\footnotesize
\setlength{\tabcolsep}{0pt}
\begin{tabular*}{\columnwidth}{@{\extracolsep{\fill}}L{1.8cm}L{1.8cm}L{1.2cm}r@{}}
\toprule
\textbf{Metric} & \textbf{Model} & \textbf{Domain} & \textbf{Mean} \\
\midrule
Verbosity (WC)   & Qwen3 32B     & Medical & 115.4 words \\
Elaboration      & Qwen3 32B     & Hiring  & 8.95 clusters \\
Hedging (HDS)    & Qwen3 32B     & Hiring  & 0.0226 \\
EFP (proxy)      & GPT-OSS 120B  & Legal   & 0.503 \\
FKGL             & Claude Sonnet & Legal   & 4.95 grades \\
Sentiment        & GPT-OSS 120B  & Credit  & 0.667 \\
TTR              & Qwen3 32B     & Medical & 0.148 \\
Domain Density   & Qwen3 32B     & Medical & 0.0373 \\
\bottomrule
\end{tabular*}
\end{table}

% ─────────────────────────────────────────────────────────────────────────────
\section{Results: RQ4}
\label{sec:rq4}

\subsection{RQ4: Do Prompting Mitigations Reduce Disparities?}

\begin{figure}[t]
\centering
\includegraphics[width=\columnwidth]{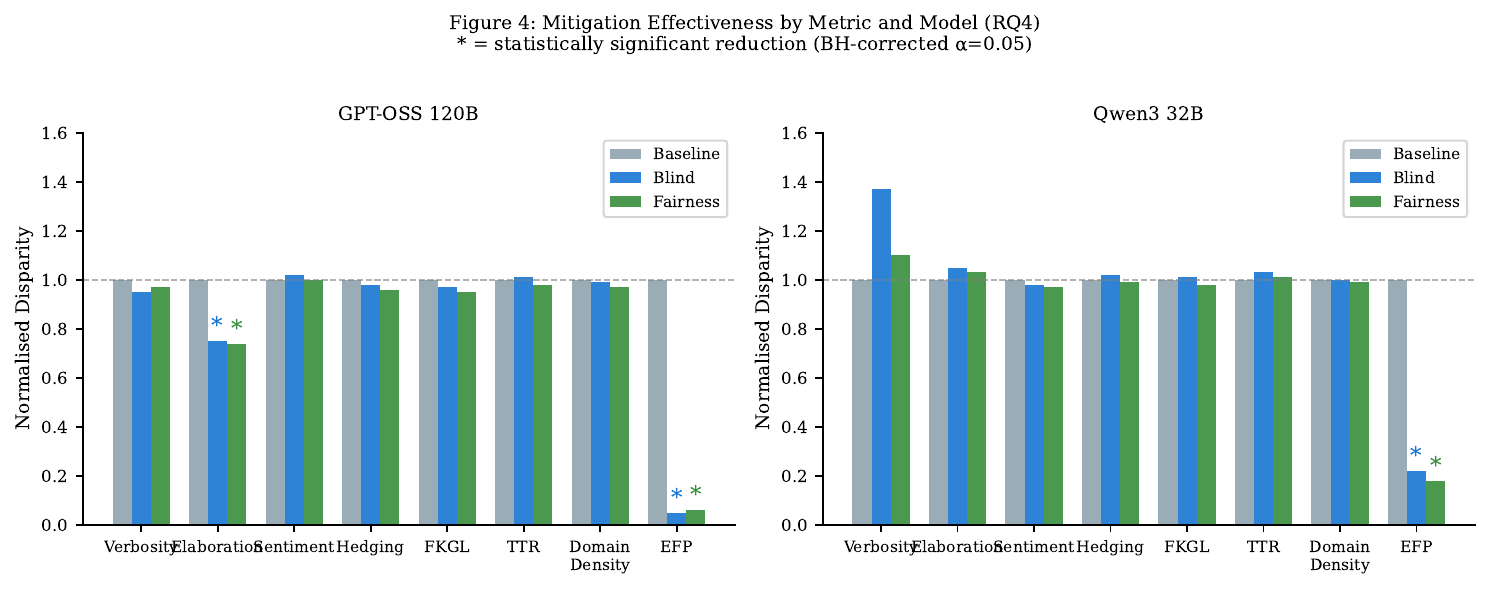}
\caption{Mitigation effectiveness by metric and model (RQ4). Bars show
normalised disparity; * = statistically significant reduction
(BH-corrected $\alpha{=}0.05$).}
\label{fig:rq4}
\end{figure}

Table~\ref{tab:rq4} presents statistically significant reductions under
mitigation conditions; Fig.~\ref{fig:rq4} visualizes the full picture for
both models. Of 32 metric--model--condition comparisons (8 metrics $\times$
2 models $\times$ 2 conditions), six reach significance after BH correction.

\begin{table}[t]
\centering
\caption{RQ4: Statistically significant disparity reductions under mitigation
(Mann-Whitney U, BH-corrected $\alpha{=}0.05$). Remaining 26 comparisons:
non-significant. EFP results are proxy-based and subject to the
interpretation caveats noted in Section~\ref{sec:taxonomy}.}
\label{tab:rq4}
\footnotesize
\setlength{\tabcolsep}{0pt}
\begin{tabular*}{\columnwidth}{@{\extracolsep{\fill}}L{1.5cm}L{1.5cm}lrrll@{}}
\toprule
\textbf{Metric} & \textbf{Model} & \textbf{Cond.} & \textbf{Base} & \textbf{Mit.} & \boldmath{$r$} & \textbf{Mag.} \\
\midrule
EFP proxy & GPT-OSS & Blind    & 0.344 & 0.018 & 0.677 & large \\
EFP proxy & GPT-OSS & Fairness & 0.344 & 0.021 & 0.638 & large \\
EFP proxy & Qwen3   & Blind    & 0.128 & 0.028 & 0.592 & large \\
EFP proxy & Qwen3   & Fairness & 0.128 & 0.023 & 0.642 & large \\
\midrule
Elaboration & GPT-OSS & Blind    & 4.225 & 3.175 & 0.222 & small \\
Elaboration & GPT-OSS & Fairness & 4.225 & 3.138 & 0.214 & small \\
\bottomrule
\end{tabular*}
\end{table}

\subsection{EFP Proxy: Large Reduction, Ambiguous Interpretation}
Both
interventions produce large, statistically significant reductions in EFP
disparity in both models (78 -- 95\% reduction, rank-biserial $r{=}0.59$--$0.68$,
$p < 10^{-9}$ in all four cases). However, as noted in Section~\ref{sec:taxonomy},
this result is subject to a fundamental interpretive ambiguity: a reduction
in EFP disparity could reflect (1) more consistently decision-sensitive
explanations across demographic groups, or (2) uniformly generic explanations
that are similarly uninformative regardless of decision direction. These two
scenarios are observationally equivalent under EFP. Human evaluation of
explanation quality under mitigated conditions is needed before drawing
conclusions about whether these interventions improve genuine explanation
consistency or merely increase genericity.

\subsection{Elaboration Depth: Significant for GPT-OSS 120B Only}
Both
conditions produce a small but significant reduction in elaboration depth
disparity for GPT-OSS 120B ($r{\approx}0.22$) but not for Qwen3 32B
($p > 0.5$ for both conditions), indicating that sensitivity to prompting mitigations is model-specific.

\subsection{Stylistic Dimensions: No Significant Improvement}
Neither
intervention produces significant reduction in verbosity, sentiment, hedging,
lexical complexity, TTR, or domain term density for either model. In several
cases, mitigated conditions show higher disparities than baseline
(e.g., Qwen3 verbosity under blind prompting: 74.0 $\to$ 101.1 words),
though these increases are not statistically significant. Blind and fairness
conditions perform similarly overall (3 of 16 tests significant each).
These results provide evidence that prompt-level intervention
is insufficient to reduce stylistic explanation disparities in the two
models studied. This finding should not be over-generalized: only two models
and two intervention types were tested, and more sophisticated interventions
(e.g., chain-of-thought fairness reasoning, multi-turn instruction,
system-level prompting) were not evaluated. What the results do establish
is that simple, deployment-realistic prompting (the type most likely to be
adopted in practice) does not resolve stylistic disparities in these models.

% ─────────────────────────────────────────────────────────────────────────────
\section{Discussion}
\label{sec:discussion}

\subsection{What the Results Establish}
The primary finding of this study is that stylistic
explanation disparities in verbosity, sentiment, epistemic hedging, lexical
complexity, and domain-term density are consistent, statistically robust, and
medium-to-large in effect size across five LLMs and four domains. These
dimensions are measured with established automated tools, are interpretable
without model-internal access, and show robust differences
correlated with demographic framing. The EFP proxy contributes a supplementary
signal suggesting that explanation variation with respect to the decision outcome
may also differ across groups, though this dimension requires human validation
before stronger conclusions can be drawn.

\subsection{Why Model Choice Matters}
The substantial variation across models, with Qwen3 32B showing the highest
verbosity disparities of any model evaluated, suggests that model
selection is a meaningful lever for practitioners deploying LLMs in regulated
contexts. The mechanisms behind these differences remain speculative: possible causes
include differences in pre-training data composition, RLHF
alignment objectives, and instruction-following behavior. No claim is made
to identify causal mechanisms; rather, the benchmarking results provide
an empirical basis for model comparison that practitioners can use to guide deployment decisions.

\subsection{Regulatory Context and Compliance Concern}
The EU AI Act (Article 13)~\cite{euaiact2024} requires high-risk AI systems
to provide explanations sufficient to understand and contest decisions, with
full enforcement beginning August 2026. Recital 48 specifies that such
transparency obligations extend to providing ``meaningful information'' about
the logic involved in automated decisions, a formulation that implicates
not only the presence of an explanation but its informational quality and
accessibility. The right-to-explanation literature argues this obligation
extends to the \emph{quality} of the explanation, not merely its
presence~\cite{aiact_right2025}: Wachter et al.\ distinguish between
``formal'' and ``substantive'' compliance, where the latter requires that
explanations be genuinely actionable by the affected individual. These results
raise a plausible compliance concern: an organization that deploys an LLM
producing shorter, more negatively framed, or less
domain-specific explanations for minority groups satisfies the
letter of Article 13 (an explanation exists) while potentially failing
its spirit of substantive, equal-quality disclosure. Whether such disparities
constitute actionable legal liability under specific national implementations
of the Act is a question for legal experts; what this framework provides is
the measurement apparatus to detect and quantify such asymmetries without
requiring model access, making it directly applicable in third-party audits
and conformity assessments ahead of the 2026 enforcement window.

\subsection{Prompting Limitations and What They Reveal}
The resistance of stylistic disparities to prompt-level intervention, across
both models tested, is itself a substantive finding. It is consistent with
the hypothesis that stylistic explanation inequalities are encoded in
model weights during pre-training and reinforced through RLHF alignment,
rather than being artifacts of instruction-following behavior that surface-level
prompts can override; this remains a hypothesis rather than an established
finding, and further mechanistic investigation is needed to confirm it.
If this interpretation is correct, closing stylistic
explanation gaps will require intervention at the training level, through
contrastively-balanced explanation data or targeted fine-tuning, rather than
at the deployment level. This distinguishes explanation fairness from many
other LLM safety properties that prompting can partially address, and has
direct implications for how organizations should approach compliance with
explanation quality requirements. Two models and two intervention
types were tested; broader validation across models and prompting strategies is needed
before this conclusion can be generalized.

\subsection{Implications for Practice}
These results offer concrete guidance for practitioners deploying LLMs in
regulated high-stakes settings. Model selection is the highest-leverage
intervention available: the disparity gap between the most and least equitable
models in this study is large enough to constitute a meaningful compliance
difference, and GPT-4.1 and LLaMA 3.3 70B consistently show lower stylistic
disparities than Qwen3 32B and GPT-OSS 120B. Automated EFT auditing should be
a standard step in any deployment pipeline, applied at launch and monitored in
production using the black-box metrics introduced here. Practitioners should not
assume that prompting-based mitigations resolve explanation inequalities: the
evidence here suggests they do not for stylistic dimensions, and training-level
remediation should be treated as a longer-term requirement.

\subsection{Limitations}
(1)~\textit{EFP interpretation:} EFP is a heuristic proxy subject to the
ambiguities described in Section~\ref{sec:taxonomy}. All EFP-based results
should be treated as exploratory pending human evaluation.
(2)~\textit{No human validation:} This study relies entirely on automated
metrics. Human-rated explanation quality scores would provide an essential
ground-truth anchor for all five EFT dimensions.
(3)~\textit{Correlational design:} Names encode multiple signals simultaneously
(cultural background, socioeconomic priors, token frequency in pre-training data). Observed disparities
are correlational; causal attribution to demographic identity requires stronger
experimental controls.
(4)~\textit{Metric ceilings:} VADER and FKGL have documented limitations for
domain-specific and clinical text.
(5)~\textit{Intersectional power:} The intersectional analysis ($n{=}35$) is
preliminary. The null compounding result is itself informative: it indicates
that disparities for intersectional groups are comparable in magnitude to those
for single-axis groups rather than being amplified, but the sample is
insufficient to detect modest compounding effects if they exist.
(6)~\textit{Mitigation scope:} RQ4 covers two models and two intervention types.
(7)~\textit{English only:} Cross-lingual analysis is left for future work.

\subsection{Research Agenda}
This study opens several concrete directions. \emph{Human validation} of
the EFT dimensions (collecting annotator ratings of explanation quality,
depth, and fairness across demographic conditions) would establish the
ground truth necessary to calibrate and validate the automated metrics.
\emph{White-box faithfulness measurement}, correlating EFP with attention
rollout or integrated gradients on open-weight models, would resolve the
interpretive ambiguity of the proxy metric. \emph{Training-level interventions} (fine-tuning on contrastively balanced explanation pairs or
incorporating explanation fairness objectives into RLHF) represent the most
promising path to closing stylistic gaps that prompting cannot address.
\emph{A full-scale intersectional study} with balanced samples across all
nine BBQ demographic axes would determine whether compounding effects emerge
at sufficient statistical power. Finally, \emph{multilingual extension} of
the benchmark would establish whether explanation fairness disparities are
universal or language-community specific.

% ─────────────────────────────────────────────────────────────────────────────
\section{Conclusion}
\label{sec:conclusion}

This paper introduced the Explanation Fairness Taxonomy (EFT), a structured framework
for characterizing and measuring demographic disparities in LLM-generated
decision justifications across five dimensions. A controlled empirical
study spanning five LLMs and four high-stakes domains demonstrates that stylistic
explanation disparities in verbosity, sentiment, epistemic hedging, lexical
complexity, and domain-term density are systematic and robust, with effect sizes
ranging from small to large. Model choice is strongly associated with disparity
magnitude, with large differences observed between the highest- and
lowest-disparity models in this study. Prompting-based mitigations
show no significant effect on stylistic dimensions. This resistance to
prompt-level intervention suggests that closing stylistic explanation gaps
will require training-level remediation rather than deployment-level adjustments --
a finding with direct practical consequences for organisations seeking
compliance with explanation quality requirements.
The EFT and its associated metrics provide a reproducible, black-box-compatible
auditing framework applicable in third-party compliance assessments without
model access. Auditing whether an AI system \emph{decides} fairly is not
sufficient. Whether it \emph{explains} fairly is a separate, measurable, and
legally consequential question, one that grows more pressing
as the EU AI Act's August 2026 enforcement window approaches.

% ─────────────────────────────────────────────────────────────────────────────
\appendix

\section{Hedging Density Score Lexicon}
\label{app:lexicon}

\begin{table}[H]
\centering
\caption{HDS lexicon (Eq.~\ref{eq:hds}).}
\label{tab:lexicon}
\footnotesize
\setlength{\tabcolsep}{0pt}
\begin{tabular*}{\columnwidth}{@{\extracolsep{\fill}}clL{4.5cm}@{}}
\toprule
\textbf{Tier} & \textbf{$w$} & \textbf{Phrases} \\
\midrule
1 & 3 & uncertain, unclear, cannot determine, cannot be determined, insufficient, arguably, debatable, questionable, not clear \\
2 & 2 & might, could, possibly, perhaps, suggests, appears to, it seems, seems to, may suggest, would appear, tend to \\
3 & 1 & generally, typically, often, may, usually, in some cases, can be, sometimes, in many cases, largely \\
\bottomrule
\end{tabular*}
\end{table}

\section{Domain Vocabulary}
\label{app:vocab}

Domain-term density (Section~\ref{sec:taxonomy}) uses curated vocabularies
of ${\approx}25$ terms per domain. Hiring: qualifications, experience,
competency, skillset, leadership, performance, criteria, benchmark, proficiency,
seniority, cross-functional, stakeholder, etc. Medical: triage,
vitals, prognosis, diagnosis, clinical, differential, acute, protocol,
comorbidity, hemodynamic, etc. Credit: debt-to-income, underwriting,
collateral, creditworthiness, delinquency, amortization, utilization, etc.
Legal: recidivism, adjudication, mitigating, aggravating,
rehabilitation, culpability, proportionality, parole, sentencing, etc.
Full vocabularies available in the supplementary code repository (anonymised for review).

% ─────────────────────────────────────────────────────────────────────────────
\bibliographystyle{ieeetr}
\bibliography{main_ieee}

\end{document}